\pdfoutput=1

\documentclass[11pt]{article}

\usepackage{arxiv}
\usepackage[numbers, sort&compress]{natbib}

\usepackage[table]{xcolor}

\usepackage{xcolor}
\usepackage[most]{tcolorbox}
\usepackage[framemethod=tikz]{mdframed}
\usepackage{xparse}
\usepackage{tikz}

\NewDocumentCommand{\boxtxt}{m m}{%
\begin{tcolorbox}[
    enhanced,
    colback=blue!5!white,
    colframe=blue!75!black,
    boxrule=1.0pt,
    arc=3mm, left=2mm, right=2mm,
    top=3mm,
    attach boxed title to top left={
        xshift=3mm, yshift=-3mm
    },
    boxed title style={
        colback=blue!25!white,
        colframe=blue!75!black,
        boxrule=1.0pt,
        left=1mm, right=1mm, top=0mm, bottom=0mm
    },
    fonttitle=\bfseries,
    coltitle=black,
    title={#1}
  ]
  \textit{#2}
\end{tcolorbox}%
}

\newtcbox{\bracketeq}{
  nobeforeafter, enhanced, colback=gray!10, frame empty, math upper,
  boxsep=2pt, left=8pt, right=8pt, top=4pt, bottom=4pt, arc=4pt,
  tcbox raise base, 
  overlay={
    \def\thk{1.0pt}   
    \def\stub{4pt}    
    \draw[line width=\thk] (frame.north west) -- (frame.south west);
    \draw[line width=\thk] (frame.north east) -- (frame.south east);
    \draw[line width=\thk] (frame.north west) -- ([xshift=\stub]frame.north west);
    \draw[line width=\thk] (frame.south west) -- ([xshift=\stub]frame.south west);
    \draw[line width=\thk] (frame.north east) -- ([xshift=-\stub]frame.north east);
    \draw[line width=\thk] (frame.south east) -- ([xshift=-\stub]frame.south east);
  }
}

\newtcbox{\boxeq}{%
  enhanced,
  on line,
  nobeforeafter,
  math upper,          
  frame empty,
  colback=gray!10,
  boxsep=2pt, left=4pt, right=4pt, top=4pt, bottom=4pt,
  arc=3pt,
  tcbox raise base,
  overlay={
    \def\r{3pt}         
    \def\thk{1.0pt}     
    \def\stub{4pt}      
    \def\wipe{1pt}      

    \draw[line width=\thk, rounded corners=\r]
      (frame.north west) rectangle (frame.south east);

    \fill[white]
      ([xshift=\stub,yshift=\wipe]frame.north west)
      rectangle
      ([xshift=-\stub,yshift=-\wipe]frame.north east);
    \fill[white]
      ([xshift=\stub,yshift=\wipe]frame.south west)
      rectangle
      ([xshift=-\stub,yshift=-\wipe]frame.south east);
  }
}

\usepackage[utf8]{inputenc} 
\usepackage[T1]{fontenc}    
\usepackage{hyperref}       
\usepackage{url}            
\usepackage{booktabs}       
\usepackage{amsfonts}       
\usepackage{nicefrac}       
\usepackage{microtype}      

\usepackage{multicol}
\usepackage{booktabs}
\usepackage{multirow}
\usepackage{threeparttable}
\usepackage{float}
\usepackage{amsmath}
\usepackage{enumitem}
\usepackage{wrapfig}
\usepackage{hhline}

\usepackage{subcaption}

\usepackage{graphicx}
\usepackage{bm}
\usepackage{bbm}

\definecolor{cafmblue}{HTML}{1D5B79}
\definecolor{cafmbluebg}{HTML}{EAF3F7}

\hypersetup{
  colorlinks=true,
  linkcolor=cafmblue,
  citecolor=cafmblue,
  urlcolor=cafmblue
}

\newcommand{\shad}{\cellcolor{cafmbluebg}} 
\newcommand{\vbench}[1]{\rotatebox[origin=c]{90}{#1}}
\newcommand{\jacob}[1]{\textcolor{blue}{[Jacob: #1]}}

\newtheorem{theorem}{Theorem}[section]

\title{Constraint-Aware Flow Matching: Decision Aligned\\ End-to-End Training for Constrained Sampling}

\author{
  Jacob K. Christopher \\
  University of Virginia \\
  \texttt{csk4sr@virginia.edu} \\
  \And
  James E. Warner \\
  NASA Langley Research Center \\
  \texttt{james.e.warner@nasa.gov} \\
  \And
  Ferdinando Fioretto \\
  University of Virginia \\
  \texttt{fioretto@virginia.edu} \\
}

\begin{document}

\maketitle

\begin{abstract}
  Deep generative models provide state-of-the-art performance across a wide array of applications, with recent studies showing increasing applicability for science and engineering. Despite a growing corpus of literature focused on the integration of physics-based constraints into the generation process, existing approaches fail to enforce strict constraint satisfaction while maintaining sample quality. In particular, training-free constrained sampling methods, while providing per-sample feasibility guarantees, introduce a fundamental mismatch between the training objective and the constrained sampling procedure, often leading to performance degradation. Identifying this training-sampling misalignment as a central limitation of current constrained generative modeling approaches, this paper proposes \textit{Constraint-Aware Flow Matching}, a novel end-to-end framework that explicitly incorporates constraint projections into the training objective. By aligning the model’s learned dynamics with the constrained sampling process, the proposed method mitigates distributional shift induced by projection-based corrections, enabling high-quality constrained generation. The proposed approach is evaluated on three challenging real-world benchmarks, illustrating the generality and efficacy of the method.
\end{abstract}


\section{Introduction}

Flow matching and diffusion generative models provide state-of-the-art performance across a wide array of settings, representing the forefront of content creation for image and video generation \cite{rombach2022high, ifriqi2025flowception, jin2024pyramidal}, engineering \cite{wang2023diffusebot}, material science \cite{cui2026constrained,christopher2025physics,wu2026dmflow}, and other scientific applications \cite{christopher2025constrained, anand2022protein, hoogeboom2022equivariant}. Despite the promising potential of these models, the stochastic nature of these approaches often results in generations which resemble the training distribution but lack integral requirements of real-world data. Particularly within scientific domains, where samples are required to precisely satisfy domain-specific constraints and physical laws to maintain any real meaning, a critical barrier to widespread adoption is the inability of generative models to adhere to these requirements.

In response to these challenges, recent literature has started exploring constrained generative modeling for scientific domains. This work can be partitioned into three classes: (1) \textit{Constraint Guidance} approaches condition the generation on the constraint set, but these methods provide no formal certificate of feasibility, thus relying heavily on rejection sampling \cite{carvalho2023motion, maze2023diffusion, botteghi2023trajectory}; (2) \textit{Physics-Informed Training} augments generative model training with a physical residual loss, aligning the sampling distribution with the constraint set but failing to enforce per-sample satisfaction \cite{warner2025generative, bastek2024physics}; (3) \textit{Constrained Sampling} provides per-sample guarantees for constraint satisfaction, and, thus, these approaches have been most widely adopted \cite{christopher2024constrained, liang2025simultaneous, utkarsh2025physics, zampini2025training, cheng2024gradient}. However, while constrained sampling shows significant promise, generation quality often degrades as compared to the unconstrained sampling approaches. 

\begin{figure}
    \centering
    \includegraphics[width=1\linewidth]{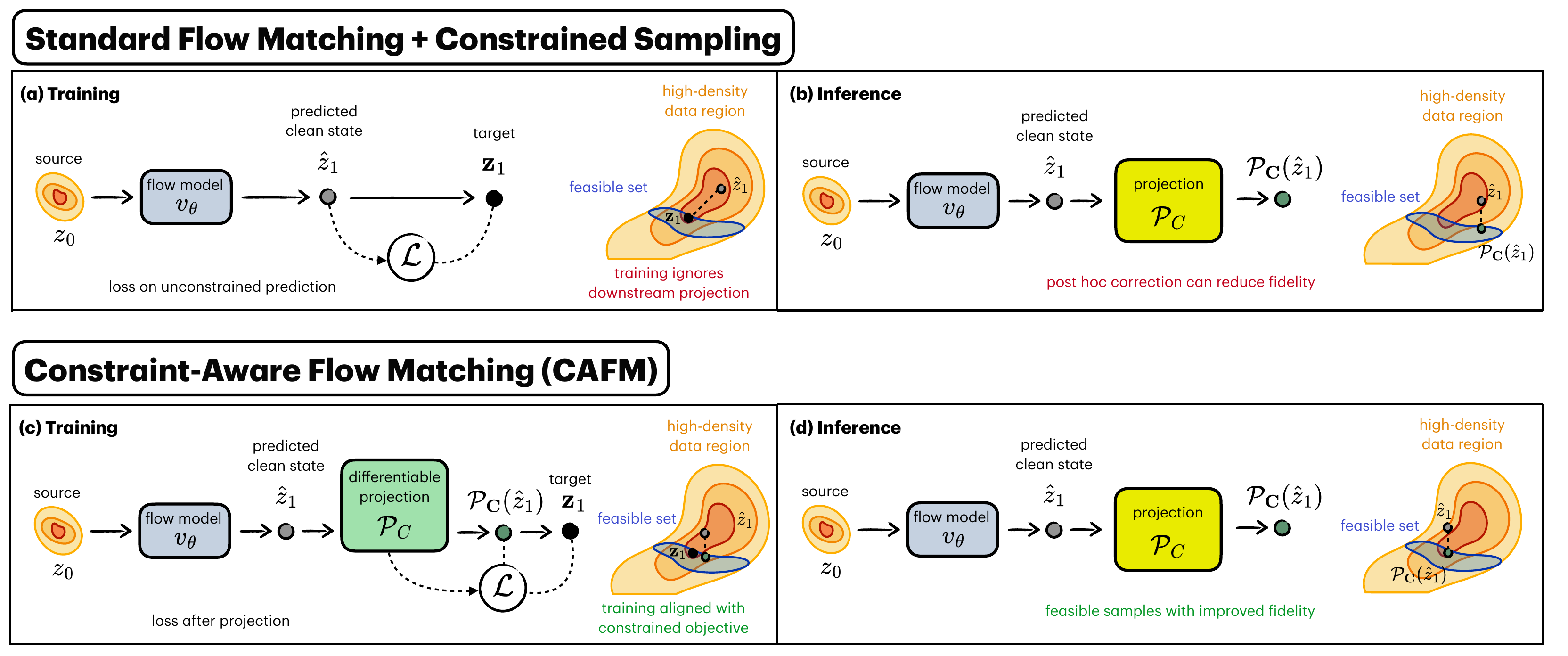}
    \vspace{-10pt}
    \caption{Visualization of \textbf{Constraint-Aware Flow Matching} compared to standard flow matching. While the clean state prediction from standard flow matching, $\hat{z}_1$, falls in a high-density region of the distribution, the projection degrades fidelity. Conversely, our constraint-aware objective optimizes the \textit{downstream task}, learning to predict $\hat{z}_1$ such that the projection falls in high-density regions.}
    \label{fig:method_visualizer}
\end{figure}

This paper argues that this performance trade-off stems from the misalignment between the training objective and the constrained sampling process.
Existing constrained sampling approaches are training-free, modifying only the sampling phase and leveraging pretrained flow matching or diffusion models. While this makes the integration of constraints inexpensive, requiring no model retraining, it also results in a model which is trained to sample in a fundamentally different way from how it is used at deployment. 
Generally, state-of-the-art methods rely on the integration of projections onto the feasible set during sampling, pushing the samples towards low density regions which may not have been learned during the training process \cite{christopher2024constrained, liang2025simultaneous, utkarsh2025physics}.
Figure \ref{fig:method_visualizer} (b) provides an illustration of this: while the initial predicted sample falls in a high-density region of the data distribution, the projection pushes the sample to lower density regions when restoring feasibility.
The paper addresses this gap by introducing a constraint-aware training procedure, ensuring alignment between training and sampling.

\textbf{Contributions.}
This work addresses the disconnect between the training and sampling processes of constrained flow matching approaches by making the following contributions: \textbf{(1)} It presents a novel end-to-end formulation of the constrained generation task, analyzing the failure modes of training-free constrained modeling (visualized in Figure \ref{fig:method_visualizer}). \textbf{(2)} It derives a constraint-aware training objective, inspired by the recent progress in the areas of differentiable optimization and decision-focused learning \cite{wilder2019melding, wang2019satnet, aaaiFerberWDT20, elmachtoub2022smart, agrawal2019differentiable, berthet2020learning, FYblondel2020, ferber2022surco, blondel2020fast, Fioretto:jair24}, 
aligning the training process with the downstream generative task and providing the first end-to-end training approach for constrained sampling approaches. 
\textbf{(3)} It presents empirical evaluation on three real-world scientific settings encompassing PDE-constrained generation, microweather forecasting, and microstructure generation, with the introduced \textit{Constraint-Aware Flow Matching (CAFM)} reporting state-of-the-art performance across these domains.

\section{Related Work}
\label{sec:related_work}

\textbf{Constraint Guidance and Physics-Informed Training.} 
Early applications of generative models for constrained domains sought to control sample dynamics using model conditioning approaches, relying on learning-based approaches which either trained an independent classifier or leveraged the diffusion model directly to steer the generation \cite{ho2020denoising, ho2022classifier}. As these methods were increasingly applied for scientific tasks \cite{wang2023diffusebot, carvalho2023motion, maze2023diffusion}, more formal notions of constraint-aware training began to be explored within the literature, applying principles from physics-informed neural networks to generative contexts \cite{raissi2019physics, raissi2018hidden}. \citeauthor{bastek2024physics} (\citeyear{bastek2024physics}) proposed incorporating a physical constraint directly into the training objective of the diffusion model. \citeauthor{warner2025generative} (\citeyear{warner2025generative}) demonstrated that physical and statistical constraints could be adopted as a residual loss when learning a latent space for latent flow matching \cite{dao2023flow}. However, while these works provide a stronger alternative to conditioning-based constraint enforcement, they fail to provide per-sample guarantees, align the sampling process only at a distributional-level, and rely heavily on neural networks to approximate the true constraints.

\textbf{Constrained Sampling.} 
More recently, \citeauthor{christopher2024constrained} (\citeyear{christopher2024constrained}) introduced constrained sampling approaches for diffusion models, incorporating constraint correction during the generation process by projecting onto the feasible set. Subsequent studies have demonstrated that constrained sampling provides state-of-the-art performance in a broad range of domains including robotics \cite{liang2025simultaneous}, biology \cite{christopher2025constrained}, and material science \cite{cui2026constrained}. Sampling-time approaches have been extended to latent diffusion models \cite{zampini2025training}, as well as to flow matching parameterizations \cite{utkarsh2025physics, liang2025chance}. However, while this class of methods contribute the strongest performance for constrained generative modeling tasks, they remain training-free by design, and this disconnect between training and sampling often results in degraded sample quality by conventional metrics.

\textbf{Differentiable Optimization.} 
Our method builds on the general principle that optimization procedures can be embedded within differentiable computational pipelines.
Differentiation through constrained optimization problems was identified as of significant importance for machine learning settings, dating back to \citeauthor{gould2016differentiating} (\citeyear{gould2016differentiating}). 
This line of research has been predominantly applied in end-to-end learning settings, enabling machine learning models to integrate constraint programming into their training processes.
\citeauthor{amos2017optnet} (\citeyear{amos2017optnet}) proposed using a quadratic programming layer for solving constrained optimization problems during the forward pass of the network. While this approach leveraged implicit differentiation of KKT conditions for the backward pass, differentiable optimization layers have been widely generalized \cite{agrawal2019differentiable, kotary2023analyzing, wilder2019melding, mandi2020interior, ferber2020mipaal, poganvcic2019differentiation, sahoo2022backpropagation, niepert2021implicit}. While many of these methods were limited to convex constraint sets, \citeauthor{kotary2023analyzing} (\citeyear{kotary2023analyzing}) and \citeauthor{blondel2022efficient} (\citeyear{blondel2022efficient}) demonstrated applicability for nonconvex constraints using sequential quadratic programming and fixed-point differentiation, enabling greater generality of applicable use cases.

\section{Preliminaries: Flow Matching}

\textbf{Flow Matching \cite{lipman2022flow}.}
Flow matching is a generative modeling framework where a neural network
learns a time-dependent velocity field 
$v_\theta : \mathbb{R}^d \times [0,1] \to \mathbb{R}^d$ 
that defines a deterministic flow $\psi_t : \mathbb{R}^d \to \mathbb{R}^d$ transporting samples from a base distribution $p_0$ to a target distribution $p_1$. Given an initial condition $z_0 \sim p_0$, the generative trajectory is obtained by solving the ODE
\begin{equation}
    \frac{d}{dt}\psi_t(z_0) \;=\; v_\theta\!\bigl(\psi_t(z_0), t\bigr),
    \qquad \psi_0(z_0)=z_0,
    \label{eq:fm_ode}
\end{equation}
and the generated sample is $z_1 = \psi_1(z_0)$.
Training in flow matching is performed using a \emph{reference transport path} constructed from paired endpoints $(\mathbf{z}_0,\mathbf{z}_1)$, where $\mathbf{z}_0\sim p_0$ and $\mathbf{z}_1\sim p_1$ are coupled (e.g., via an optimal transport plan). A time-$t$ point on this reference path is denoted $\mathbf{z}_t$ (e.g., $\mathbf{z}_t=(1-t)\mathbf{z}_0+t\mathbf{z}_1$, optionally with small noise), and the corresponding reference velocity for the path is $\mathbf{z}_1-\mathbf{z}_0$. The velocity network $v_\theta$ is then trained by regressing to this reference velocity:
\begin{equation}
    \mathcal{L}_{FM} = \|\overbrace{{v}_\theta (\mathbf{z}_t, t)}^{\text{predicted}} - \overbrace{(\mathbf{z}_1 - \mathbf{z}_0)}^{\text{true velocity}}\|^2.
    \label{eq:flow_matching}
\end{equation}

\textbf{Notation.}
Bold variables $\mathbf{z}_0,\mathbf{z}_t,\mathbf{z}_1$ denote states induced by the \emph{training} coupling/path (constructed directly from paired endpoints), whereas non-bold variables $z_0,z_t,z_1$ denote states induced by the \emph{learned generative flow} in \eqref{eq:fm_ode}. This distinction is important because training evaluates $v_\theta$ on $\mathbf{z}_t$ sampled from the reference path, while at generation time the model is queried on $z_t$ produced by integrating the learned dynamics.

\section{Constraint-Aware Sampling}
\label{sec:method}

The goal of constrained generation is to produce samples $z_1 \sim p_\text{data}$ subject to a set of feasibility requirements $C$. 
While often times the true data distribution $p_\text{data}$ may already satisfy the relevant constraints by construction (e.g., physical systems typically respect the underlying governing laws), the practical challenge is that a learned generative model $v_\theta$, despite approximating $p_\text{data}$, may assign probability mass to infeasible regions of the ambient space.
If $\mathbbm{1}(\cdot)$ denotes an indicator function, the \emph{idealized constrained target} is:
\begin{align}
    p_{C}(z_1)  \propto p_\text{data}(z_1) \mathbbm{1}\{z_1 \in C\},
    \label{eq:gen_obj}
\end{align}

with the practical objective obtained by replacing $p_\text{data}$ with the learned approximation obtained by $v_\theta$ \cite{christopher2025constrained}. Across applications, the constraint set $C$ is selected on the basis of domain-specific criteria, but, to ensure generality in the presentation, it can be arbitrarily defined as the intersection of a series of equality constraints $\mathbf{h}(z_1) =0$ and inequality constraints $\mathbf{g}(z_1) \leq 0$, imposed on the ambient space.

As discussed in Section \ref{sec:related_work}, constrained sampling approaches provide state-of-the-art performance for this class of generation task. 
However, these methods suffer from their own set of challenges. Seminal works focus on correcting intermediate or partially denoised states to allow constraint enforcement to guide the generation process \cite{christopher2024constrained, liang2025simultaneous}. 
Yet, enforcing constraints on noisy representations can bias the sampling trajectory, distort the learned dynamics, or over-correct due to high-variance noise \cite{blanke2025strictly}. More recent studies have demonstrated that enforcing feasibility on clean state predictions results in a reduction of this bias, particularly in high-dimensional settings within more restrictive feasibility criteria \cite{utkarsh2025physics, christopher2025constrained}. In consequence, this study adopts a final state projection method formalized below:
\begin{align}
    \hat{z}_1 &= \textrm{ODESolve}\bigl( z_t, v_\theta, t, 1\bigr) \tag{\text{Forward Solve}}  \label{eq:fwd} \\
    z_1 &= \mathcal{P}_C \bigl(\hat{z}_1\bigr) \tag{\text{Constraint Projection}} \label{eq:proj} \\
    z_{s} &= \textrm{ODESolve}\Bigl(z_1, - \bigl(z_1 - z_0 \bigr), 1, s\Bigl) \tag{\text{Reverse Update}}
\end{align}
where the projection operator $\mathcal{P}_C(y) := \arg\min_{x \in C} \|x - y\|^2_2$ returns the nearest feasible point, the $\textrm{ODESolve}$ follows the trajectory defined in Equation \eqref{eq:fm_ode}, and timestep $s = t +\Delta$ for a step size $\Delta > 0$.
This formulation follows the sampling approach proposed by \cite{utkarsh2025physics}, which removes the intermediate feasibility assumption presented in prior work, solely enforcing \textit{terminal state feasibility}. At each sampling stage, the \textit{Forward Solve} predicts a clean $z_1$, which is projected onto the constraint manifold. Then, the learned flow is updated such that the end point enforces feasibility.


\section{Aligning Training and Inference}

Within the existing constrained sampling literature, the training-free property of the sampling approaches is often regarded as a merit of the methodology. 
However, 
existing diffusion and flow matching objectives are designed to sample from $p_\text{data}$, thus remaining misaligned with the downstream task of sampling from $p_C$. 
Indeed, the standard training objective optimizes the unconstrained generative path (Figure \ref{fig:method_visualizer} (a)), while the actual deployment objective is strongly influenced by the feasibility correction (Figure \ref{fig:method_visualizer} (b)). These observations suggest that sampling-time enforcement alone creates a disconnect between what the models is trained to optimize and what is ultimately evaluated in deployment.

While this challenge remains unaddressed for constrained generative modeling, machine learning for optimization is an area of the literature where end-to-end learning has been explored extensively \cite{mandi2024decision}. This area of research formalizes a downstream task:
\begin{subequations}
\label{eq:dfl_problem}
\begin{align}
    \mathbf{x}^\star(\mathbf{c}) = \underset{\mathbf{x}}{\arg\min} \; &f(\mathbf{x}, \mathbf{c}) \\
    \textrm{s.t.} \;\;\; &\mathbf{g}(\mathbf{x}, \mathbf{c}) \leq 0; \quad
    \mathbf{h}(\mathbf{x}, \mathbf{c}) = 0
\end{align}
\end{subequations}
where $\mathbf{x}^\star(\mathbf{c})$ is an optimal decision of a constrained optimization problem parameterized by $\mathbf{c}$, a prediction by an ML model. While earlier works focused on prediction-focused learning \cite{hu2016toward}, minimizing the mean squared error between the predicted parameters $\mathbf{\hat{c}}$ and the true parameters $\mathbf{c}$, later studies showed that decision-focused learning \cite{wilder2019melding}, minimizing the downstream loss,
\begin{align}
    \mathcal{L}_{\text{DFL}} := f(\mathbf{x}^\star(\mathbf{\hat{c}}), \mathbf{c}) -  f(\mathbf{x}^\star(\mathbf{{c}}), \mathbf{c}), 
    \label{eq:loss_dfl}
\end{align}
results in improved downstream performance. Theoretically, this improvement provided by decision-focused learning, Equation \eqref{eq:loss_dfl}, can be characterized by a discontinuity between the the downstream and the prediction-focused loss; while the global optimum between the two losses is aligned, sharp changes in $\mathbf{x}^\star(\mathbf{\hat{c}})$ can result in suboptimality increasing even when prediction quality improves \cite{demirovic2019investigation}.



\subsection{Constraint-Aware Flow Matching Objective}
Building from the problem formulation presented in Equation \eqref{eq:dfl_problem}, this section extends the end-to-end formulation leveraged by decision-focused learning to the setting of constrained generation. 
Bridging the decision-focused learning formulation to the constrained generation task in Equation \eqref{eq:gen_obj}, let $\mathbf{x}^\star(\mathbf{c}) := z_1$, $\mathbf{c} := \mathbf{z}_1$, and $f(\mathbf{x}, \mathbf{c}) := \| \mathbf{x} - \mathbf{c}\|^2$, the objective of the projection operator, recovering the optimization:
\begin{subequations}
    \label{eq:pafm_co}
    \begin{align}
        \label{eq:pafm_co_obj}
        \mathbf{x}^\star(\mathbf{c}) = \underset{\mathbf{x}}{\arg\min} \;  \| &\mathbf{x} - \mathbf{c}\|^2 \\
        \textrm{s.t.} \;\;\; &\mathbf{g}(\mathbf{x}) \leq 0; \quad
        \mathbf{h}(\mathbf{x}) = 0
    \end{align}
\end{subequations}
Assuming $\mathbf{z}_1 \in {C}$, this formulation recovers the training sample exactly, analogous to decision-focused learning under the true parameterization of $\mathbf{c}$. 
At training time, then, the standard flow matching objective aligns with prediction-focused learning; specifically, it optimizes the prediction task directly, as we define $\mathbf{\hat{c}} := \hat{z}_1$, the unconstrained prediction.

As a result, training-free constrained sampling approaches realize a gap between the prediction-focused task and the downstream decision-focused task, mirroring the challenges that have been exhaustively studied within decision-focused learning. While this discontinuity especially emerges when applying nonconvex constraints, where small changes in the
\begin{wrapfigure}[19]{r}{0.35\linewidth}
    \vspace{-8pt}
    \centering
    \includegraphics[width=\linewidth,trim={6pt 6pt 6pt 5pt},clip]{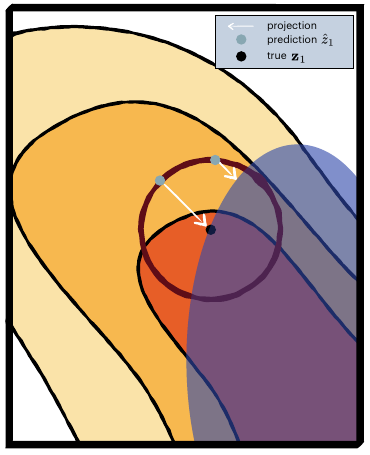}
    \vspace{-16pt}
    \renewcommand{\baselinestretch}{0.9} 
    \caption{\small Example of two predictions with identical prediction-focused losses resulting in substantially different decision-focused losses.}
    \renewcommand{\baselinestretch}{1.0} 
    \label{fig:example_convex}
\end{wrapfigure} 
prediction task can result in larges changes in the minimizer of the projection, the gap is present even in convex cases as illustrated in Figure \ref{fig:example_convex}. Indeed, the example shows that even in simple convex cases, predictions ${\hat{z}_1}$ with identical mean square error can result in very different downstream decisions $z_1$.

A natural objection could be that different parameterizations of generative models may allow training-free constrained sampling to address this gap. For instance, considering the loss landscape in Figure \ref{fig:example_convex}, one may propose adopting a proximal method to sample from high density regions while satisfying the constraint. As such, prior works have leveraged score-matching as an approach for estimating the density function gradient, enabling proximal gradient optimization \cite{christopher2024constrained, blanke2025strictly}.
However, the disconnect between the training and sampling process remains the central challenge in this case as well; as the proximal algorithm diverges from the forward diffusion trajectory, score estimation becomes increasingly inaccurate due to the absence of training data in these low density regions \cite{christopher2025constrained, song2019generative}.
Hence, a critical step to closing the gap with standard generative methods remains defining an aligned, end-to-end training process for constrained samplers.



In order to derive this end-to-end training objective, we begin by revisiting the standard flow matching loss function:
\begin{subequations}
    \begin{align*}
    \mathcal{L}_{FM} = \|\overbrace{{v}_\theta (\mathbf{z}_t, t)}^{\text{predicted}} - \overbrace{(\mathbf{z}_1 - \mathbf{z}_0)}^{\text{true velocity}}\|^2 
    &=\; 
    \| \bigl( \mathbf{z}_0 + v_\theta(\mathbf{z}_t, t) \bigr) - \mathbf{z}_1 \|^2 \\
    & = \| \hat{z}_1 -  \mathbf{z}_1  \|^2 
    \;=\; \| \mathbf{\hat{c}} -\mathbf{c} \|^2 \tag{Prediction-Focused Loss}
    \end{align*}
\end{subequations}

To align the flow matching loss with the downstream task, the mean square error optimization is imposed on the optimal solution $\mathbf{x}^\star(\mathbf{\hat{c}}):= z_1$ rather than $\mathbf{\hat{c}}:= \hat{z}_1$, restoring alignment with the true endpoint. 
Equation \eqref{eq:flow_matching} is thus extended under this framing to yield the \textbf{constraint-aware flow matching objective}:
\begin{subequations}
\label{eq:pafm_obj}
\begin{align}
    \mathcal{L}_{CAFM} &= \Bigl\|\overbrace{\Bigl(\mathcal{P}_C\bigl(\mathbf{z}_0 + v_\theta(\mathbf{z}_t,t)\bigr) - \mathbf{z}_0\Bigr)}^{\text{projected velocity prediction}} - \overbrace{\Bigl(\mathbf{z}_1 - \mathbf{z}_0\Bigr)}^{\text{true velocity}}\Bigr\|^2 \\    
    &= \Bigl\|\overbrace{\mathcal{P}_C\bigl(\mathbf{z}_0 + v_\theta(\mathbf{z}_t,t)\bigr)}^{\mathcal{P}_C(\hat{z}_1) \; = \; z_1} - \;\mathbf{z}_1 \Bigr\|^2 
\end{align}
\end{subequations}
Notably, this recovers the decision-focused loss in Equation \eqref{eq:loss_dfl},
\begin{align*}
        \mathcal{L}_{CAFM} &= \Bigl\|\overbrace{\mathcal{P}_C\bigl(\mathbf{z}_0 + v_\theta(\mathbf{z}_t,t)\bigr)}^{\mathbf{x}} - \; \overbrace{\mathbf{z}_1}^\mathbf{c} \Bigr\|^2 \\
        &= \|\mathbf{x} - \mathbf{c} \|^2 \!- 0 
        \;=\; f(\mathbf{x}^\star(\mathbf{\hat{c}}), \mathbf{c}) -  f(\mathbf{x}^\star(\mathbf{{c}}), \mathbf{c}) \tag{Decision-Focused Loss}
\end{align*}

\subsection{Differentiable Projection Operators}
\label{sec:diff_projection}

Having justified the adoption of the constraint-aware flow matching objective, the next challenge is defining the projection problem such that it can be treated as a differentiable operation. The key observation is that, when implemented by an iterative optimization algorithm, the projection can be unrolled into the computational graph. 
To facilitate end-to-end training, the operator $\mathcal{P}_C$ is integrated as a \textit{differentiable projection layer}: a parameterized mapping from the predicted clean state $\hat{z}_1$ to the projected state $z_1 = \mathcal{P}_C(\hat{z}_1)$.
However, rather than requiring a closed-form projector, $\mathcal{P}_C$ is implemented by an iterative optimization routine. Let $\Phi$ denote a single update of this optimization algorithm, starting from the initialization $\mathbf{x}^{(0)}(\mathbf{\hat{c}}) = \hat{z}_1$, and 
\[
    \mathbf{x}^{(k+1)}(\mathbf{\hat{c}}) = \Phi\bigl(\mathbf{x}^{(k)}, \mathbf{\hat{c}} \bigr), \quad k=0,\ldots, K - 1.
\]
Then, the projection layer used during training is composed of a series of iterative updates:
\[
    \mathcal{P}_C \bigl(\hat{z}_1\bigr) = \mathbf{x}^{\star}(\mathbf{\hat{c}}) \approx \mathbf{x}^{(K)}(\mathbf{\hat{c}}) = \bigl(\Phi \circ \cdots \circ \Phi\bigr)\bigl(\hat{z}_1\bigr)
\]
where \(\mathbf{x}^{\star}(\hat{\mathbf{c}})\) denotes the fixed-point of the update map, satisfying
\[
    \mathbf{x}^{\star}(\hat{\mathbf{c}})
    =
    \Phi\bigl(\mathbf{x}^{\star}, \hat{\mathbf{c}}\bigr).
\]
Because the optimization routine is run for only \(K\) iterations, the layer uses \(\mathbf{x}^{(K)}\) as a finite-step approximation to this fixed-point.
Under this view, the unrolled operations are added to the graph via the chain rule, enabling gradient flow from the downstream objective.

\boxtxt{Projecting with Sequential Quadratic Programming.}
{\emph{To accommodate constraint sets defined by general nonlinear relations, including nonconvex structure, optimization updates $\Phi$ is practically implemented using Sequential Quadratic Programming (SQP)
correction steps. In contrast to proximal gradient updates, which are often more applicable for simple convex sets, SQP addresses nonlinearity by solving a sequence of local quadratic subproblems obtained from linearized constraints and a quadratic model of the Lagrangian}
\[
    \mathcal{L}(z_1, \lambda, \nu) =  \tfrac{1}{2}\| \hat{z}_1 - {z}_1 \|^2 + \lambda^\top h(z_1) + \nu^\top g(z_1).
\]
\emph{where $\nu \geq 0$. This follows the rationale provided by \citeauthor{kotary2023analyzing} (\citeyear{kotary2023analyzing}), who use SQP for nonconvex nonlinear constrained optimization layers, where feasibility is not enforced through a simple convex proximal map.
Appendix \ref{app:backward} provides further details of the implementation.}}

\section{Experiments}

Three real-world problems are selected for the downstream evaluation of the proposed methodology:
\begin{enumerate}[label=\textbf{\arabic*.}, 
                  leftmargin=*, 
                  itemsep=0pt, 
                  parsep=0pt, 
                  topsep=0pt]
\item \textbf{Partial Differential Equation} constrained physical systems, comprising four distinct problems,
\item \textbf{Wind Velocity Field Estimation} over spatiotemporal horizons under a prescribed coherence, 
\item \textbf{Microstructure Inverse-Design} for material discovery under porosity constraints.
\end{enumerate}
For transparent evaluation, CAFM uses an identical sampling setup to Physics-Constrained Flow Matching (PCFM) \cite{utkarsh2025physics} in all settings, isolating the impact of end-to-end training. Additional constrained sampling approaches are included when applicable, and Functional Flow Matching (FFM) \cite{lipman2022flow} serves as a reference for unconstrained generative performance. The method performance is evaluated by \textit{accuracy} and \textit{feasibility} of the outputs, with domain-specific evaluation metrics selected for each task. 
Extended details are provided in Appendix \ref{app:exp_desc}. 

\begin{figure}[h]
    \centering
    \includegraphics[width=0.99\linewidth]{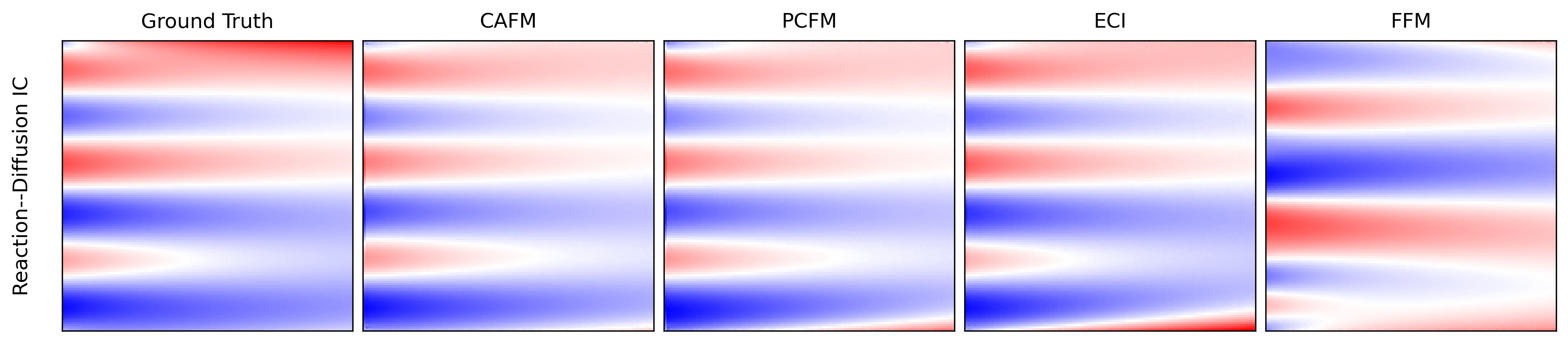}
    \caption{Visualization of baseline performance on Reaction--Diffusion IC task.}
    \label{fig:rd1d}
\end{figure}

\subsection{Application 1: Partial Differential Equations}
\label{exp:pde}

The first setting benchmarks CAFM across a set of challenging partial differential equations. Evaluations on PDE governed systems have recently been adopted as a standard stress test for constrained generative models, providing real-world scientific complexity that remains difficult for neural operators and generative models \cite{utkarsh2025physics, bastek2024physics}. 

The selected settings introduce increasing constraint complexity, as implemented by \cite{utkarsh2025physics}.
Navier--Stokes imposes initial vorticity (IC) and linear global mass conservation constraints (CL).
Reaction--Diffusion introduces curvature to the constraint manifold through nonlinear conservation law (CL). Burgers BC incorporates localized, pointwise boundary conditions increasing structural rigidity, while Burgers IC enforces initial conditions (IC) and both nonlinear global nonlinear mass conservation and local conservation updates (CL). Following \citeauthor{utkarsh2025physics} (\citeyear{utkarsh2025physics}), solution subsets are constrained on held-out IC or BC, allowing the experiment to evaluate the performance of constraint-aware training when the test-time constraints have not appeared in the training set.

\begin{table*}[t]
\vspace{-10pt}
  \centering
  \begin{threeparttable}
    {\footnotesize
    \setlength{\tabcolsep}{3.0pt}
    \renewcommand{\arraystretch}{1.22}
    \begin{tabular}{@{}llcccclcccc@{}}
      \toprule
      & & \multicolumn{2}{c}{\shad{\textit{Shared Sampler}}} & \multicolumn{2}{c}{\textit{Other Samplers}} & & \multicolumn{2}{c}{\shad{\textit{Shared Sampler}}} & \multicolumn{2}{c}{\textit{Other Samplers}} \\
      \cmidrule(lr){3-6}\cmidrule(lr){8-11}
      \textbf{Metric} & \vbench{\textbf{Benchmark}} & \shad{\textbf{CAFM}} & \shad{\textbf{PCFM}} & \textbf{ECI} & \textbf{FFM} & \vbench{\textbf{Benchmark}} & \shad{\textbf{CAFM}} & \shad{\textbf{PCFM}} & \textbf{ECI} & \textbf{FFM} \\
      \midrule
      MMSE
        & \multirow{4}{*}{\vbench{\shortstack{Reaction--\\Diffusion IC}}}
          & \shad{\textbf{1.71e-03}} & \shad{\underline{2.41e-03}} & 7.87e-02 & 4.16e-02
        & \multirow{4}{*}{\vbench{\shortstack{Navier--\\Stokes}}}
          & \shad{\underline{2.83e-01}} & \shad{2.88e-01} & 3.20e-01 & \textbf{1.98e-01} \\
      SMSE
        &
          & \shad{\textbf{1.01e-02}} & \shad{\underline{1.91e-02}} & 8.71e-02 & 3.28e-02
        &
          & \shad{1.51e-01} & \shad{1.90e-01} & \underline{5.49e-02} & \textbf{4.22e-02} \\
      CV (IC)
        &
          & \shad{\underline{1.09e-27}} & \shad{\textbf{1.01e-27}} & 6.89e-16 & 1.76e-01
        &
          & \shad{\textbf{0.00e+00}} & \shad{\textbf{0.00e+00}} & \underline{8.24e-17} & 2.12e-01 \\
      CV (CL)
        &
          & \shad{\textbf{2.22e-15}} & \shad{\underline{2.26e-15}} & 1.46e-14 & 8.08e-05
        &
          & \shad{\underline{1.26e-16}} & \shad{1.49e-16} & \textbf{5.21e-17} & 1.18e-02 \\
      \midrule
      MMSE
        & \multirow{4}{*}{\vbench{\shortstack{Burgers\\BC}}}
          & \shad{\underline{2.61e-03}} & \shad{\textbf{1.78e-03}} & 3.75e-03 & 1.16e-02
        & \multirow{4}{*}{\vbench{\shortstack{Burgers\\IC}}}
          & \shad{\textbf{3.56e-02}} & \shad{\underline{3.81e-02}} & 7.78e-02 & 4.95e-02 \\
      SMSE
        &
          & \shad{\textbf{4.55e-04}} & \shad{\underline{5.38e-04}} & 1.06e-02 & 1.58e-02
        &
          & \shad{6.85e-02} & \shad{1.05e-01} & \textbf{1.73e-02} & \underline{4.04e-02} \\
      CV (BC/IC)
        &
          & \shad{\textbf{4.44e-16}} & \shad{\textbf{4.44e-16}} & \underline{4.48e-16} & 5.42e-02
        &
          & \shad{\textbf{3.43e-22}} & \shad{\underline{3.79e-22}} & 4.31e-17 & 8.14e-02 \\
      CV (CL)
        &
          & \shad{\textbf{3.16e-15}} & \shad{\underline{3.45e-15}} & 1.16e-14 & 2.59e-04
        &
          & \shad{\textbf{3.28e-15}} & \shad{\underline{3.49e-15}} & 7.87e-12 & 1.57e-03 \\
      \bottomrule
    \end{tabular}
    }
  \end{threeparttable}
    \caption{Generative performance for zero-shot methods on constrained PDEs with linear and nonlinear constraints. 
    Navier--Stokes enforces global conservation laws (CL) as linear constraints, along with initial condition (IC) constraints. In contrast, Burgers and Reaction--Diffusion apply CL as nonlinear constraints along with IC or boundary conditions (BC) constraints. 
    Lower values indicate better performance, with best results in \textbf{bold} and second best \underline{underlined}. 
    \vspace{-8pt}
    }
    \label{tab:pde_results}
\end{table*}

Several clear trends emerge across the settings reported in Table \ref{tab:pde_results}. First, while FFM \cite{kerrigan2023functional} reports strong reconstruction metrics in applications like Navier--Stokes, the high constraint residuals demonstrate that unconstrained approaches are not viable for these complex PDE settings. Next, while the ECI framework \cite{cheng2024gradient} is generally competetive with the other constrained samplers for constraint satisfaction, it generally reports less competitive accuracy metrics as compared to the other constrained approaches. Finally, and most notably, the table shows that CAFM consistently outperforms its constrained sampling counterpart PCFM, illustrating that the end-to-end training improves both accuracy and feasibility (Figure \ref{fig:rd1d}). These results offer strong evidence that CAFM can effectively scale across increasingly difficult constraint sets. Furthermore, the strong performance illustrates that this method provides strong generalization to out-of-distribution constraint sets, given the evaluation is conducted over unseen initial conditions and boundary conditions.

\subsection{Application 2: Microweather Wind Velocity Field Estimation}
\label{exp:wind}


The next evaluation assesses the proposed method for wind velocity field prediction. Motivated by the real-world gap in prediction of microweather, localized weather conditions which are computationally prohibitive to explicitly compute with fluid dynamics, this experiment requires that the generative model predicts mean and variance wind fields while maintaining realistic spatiotemporal coherence \cite{11068624, drones8040147}.
Due to the prohibitive cost of collecting this wind velocity data, which is dependent on the deployment of sensors, the training set is composed of sparse synthetic measurements to mimic the challenges of collecting spatially-dense wind velocity data with deployed sensors.


The evaluation targets a fixed-grid version of the wind velocity field estimation setting described by \citeauthor{warner2025generative} (\citeyear{warner2025generative}), where the goal is to generate realistic wind samples from sparse spatiotemporal observations while preserving a prescribed coherence structure. As in the original setup, wind velocity is modeled using the standard decomposition \cite{carassale2006monte}:
\begin{align*}
    \mathbf{U}(\mathbf{r}, \omega) = \boldsymbol{\mu}_u(\mathbf{r}) + \mathbf{W}(\mathbf{r}, \omega)
\end{align*}
where $\mathbf{r} = [x_1, x_2, x_3, t]$ represents a static spatiotemporal point, $\boldsymbol{\mu}_u$ is a deterministic mean function over $u \in \mathbb{R}^{x_1 \times x_2 \times x_3 \times t}$, and $\mathbf{W}(\mathbf{r}, \omega)$ is a zero-mean, stationary Gaussian process.
The application considers a spatiotemporal correlation constraint, which is defined between two grid locations $\mathbf{r}$ and $\mathbf{r}'$ by the coherence function:
\begin{align*}
    \mathbf{Coh}(\mathbf{r}, \mathbf{r}', n) = \exp \Bigl( - \frac{n \| \mathbf{d}^{T}(\mathbf{r} - \mathbf{r}') \|}{\|\boldsymbol{\mu}_u(\mathbf{r}) - \boldsymbol{\mu}_u(\mathbf{r}')\|} \Bigr)    
\end{align*}
where $n$ is frequency and $\mathbf{d}$ contains directional decay coefficients. 
Following \citeauthor{warner2025generative} (\citeyear{warner2025generative}), the evaluation focuses on a two-dimensional spatial plane over time, 
$\mathbf{r} = [x_2, x_3, t]$. 
In contrast to the continuous functional representation used in the original setup, our implementation operates on a fixed grid with $N_{x_2}=10$, $N_{x_3}=10$, and $N_t=256$, so each wind sample is represented as an array
$u \in \mathbb{R}^{N_{x_2} \times N_{x_3} \times N_t} = \mathbb{R}^{10 \times 10 \times 256}$.

Empirical results are reported in Table \ref{tab:generative-performance-wind}. The unconstrained baseline, Functional Flow Matching (FFM) \cite{kerrigan2023functional}, is ineffective in this experimental setting, reporting the worst performance across all metrics; the higher MMSE and Variance MSE suggests that poor adherence to the coherence constraint results in wind velocities which fail to align with the training data. Conversely, the Projected Diffusion Model (PDM) \cite{christopher2024constrained}, which we extend to operate with the flow matching backbone as in \cite{utkarsh2025physics}, and the PCFM model report the strongest results for Variance MSE, while improving the MMSE and CV metrics over FFM. In contrast, CAFM reports the strongest MMSE and constraint violation, lowering the coherence residual by an order of magnitude. While the variance trade-off is interesting to note, as it suggests that the constraint-aware objective may suppress high-variance structures to avoid projection instability, the significant improvements in feasibility and reconstruction accuracy mark a pronounced improvement over existing constrained sampling approaches.

\begin{table}[t]
\centering

\begin{subtable}[t]{0.48\textwidth}
  \centering
  \begin{threeparttable}
    {\scriptsize
    \setlength{\tabcolsep}{7.0pt}
    \renewcommand{\arraystretch}{1.5}
    \begin{tabular}{@{}lcccc@{}}
      \toprule
      & \multicolumn{2}{c}{\shad{\textit{Shared Sampler}}} & \multicolumn{2}{c}{\textit{Other Samplers}} \\
      \cmidrule(lr){2-3}\cmidrule(lr){4-5}
      \textbf{Metric} & \shad{\textbf{CAFM (Ours)}} & \shad{\textbf{PCFM}} & \textbf{PDM} & \textbf{FFM} \\
      \midrule
      MMSE & \shad{\textbf{4.60e-02}} & \shad{\underline{4.85e-02}} & \underline{}{4.85e-02} & 5.05e-02 \\
      Variance MSE & \shad{4.73e-01} & \shad{\underline{4.17e-01}} & \textbf{4.13e-01} & 4.85e-01 \\
      CV (Coherence) & \shad{\textbf{3.60e-04}} & \shad{4.59e-03} & \underline{2.68e-03} & 5.136e-01 \\
      \bottomrule
    \end{tabular}
    }
  \end{threeparttable}
  \caption{Fixed-grid wind modeling.}
  \label{tab:generative-performance-wind}
\end{subtable}
\hfill
\begin{subtable}[t]{0.48\textwidth}
  \centering
  \begin{threeparttable}
    {\scriptsize
    \setlength{\tabcolsep}{7.0pt}
    \renewcommand{\arraystretch}{1.5}
    \begin{tabular}{@{}lcccc@{}}
      \toprule
      & \multicolumn{2}{c}{\shad{\textit{Shared Sampler}}} & \multicolumn{2}{c}{\textit{Other Samplers}} \\
      \cmidrule(lr){2-3}\cmidrule(lr){4-5}
      \textbf{Metric} & \shad{\textbf{CAFM (Ours)}} & \shad{\textbf{PCFM}} & \textbf{PDM} & \textbf{FM} \\
      \midrule
      MSE & \shad{\textbf{3.69e-01}} & \shad{5.00e-01} & \underline{4.32e-01} & 4.51e-01 \\
      NNMSE & \shad{\textbf{9.05e-02}} & \shad{\underline{9.59e-02}} & 1.41e-01 & 1.45e-01 \\
      CV (Porosity) & \shad{\textbf{0.00e+00}} & \shad{\textbf{0.00e+00}} & \textbf{0.00e+00} & 9.38e-01 \\
      \bottomrule
    \end{tabular}
    }
  \end{threeparttable}
  \caption{Microstructure inverse design.}
  \label{tab:generative-performance-micro}
\end{subtable}
\vspace{6pt}
\caption{Final metric summaries for the two generative modeling settings. Lower values indicate better performance, with best results in \textbf{bold} and second-best results \underline{underlined}.}
\label{tab:generative-performance}
\vspace{-10pt}
\end{table}



\subsection{Application 3: Microstructure Inverse-Design}
\label{exp:micro}


The final setting benchmarks the proposed method on a material science inverse-design task. Due to the prohibitive cost of collecting microstructure imaging data, generative models have been
viewed as a highly promising direction for obtaining necessary samples for discovering structure-property linkages \cite{christopher2024constrained,zampini2025training,chun2020deep}. 
For this evaluation, a sparse dataset of Bentheimer sandstone imaging data is adopted \cite{li2022bentheimer} and subsampled to create $256 \times 256$ patches rescaled to $[-1, 1]$.

This experiment adopts a convex constraint set, following the definition provided by \cite{zampini2025training}. Specifically, let $\mathbf{r}_{i,j}$ be the pixel value for row $i$ and column $j$, where $\mathbf{r}_{i,j} \in [-1, 1]$ for all values of $i$ and $j$. The porosity constraint is then,
\[
    \textit{porosity} = \tfrac{1}{n \times m}\sum^n_{i=1} \sum^m_{j=1} \mathbbm{1}{\left( \mathbf{r}_{i,j} < 0 \right)},
\]
where our constraint enforces a strict target porosity value for each sample (e.g. $\textit{porosity} = 0.5$).

The downstream evaluation metrics are reported in Table \ref{tab:generative-performance-micro}. 
The results for this task parallel the previous two settings in the general trends: FM struggles to adhere to the prescribed constraint, and the performance suffers because of this; PDM and PCFM remain competitive with one another, consistently improving over the FM baseline. Meanwhile, CAFM outperforms both PCFM and all other methods across accuracy metrics, while maintaining perfect constraint satisfaction. The empirical evaluation for this setting reiterates the benefit of aligning model training with the downstream objective, providing state-of-the-art performance when utilizing this end-to-end training paradigm.




\section{Discussion}
\label{sec:discussion}

\textbf{Connection to Multistage Optimization.}
While classical decision-focused learning formulations consider a single downstream optimization problem, many real-world systems are more accurately framed as sequential or multistage decision-making, where predictions inform a sequence of optimization steps rather than a single decision. Recent work has targeted such applications, leveraging a predictive model which, rather than parameterizing a single optimization problem, induces a policy through repeated applications of an optimization operator over a temporal horizon \cite{pervsak2024decision}.
Formally, let $\mathbf{x}_{(k)}$ denote the state at iteration $k$, and let $\mathbf{\hat{c}}_{(k)}$ denote the predicted parameters at that stage. A sequential decision process can be written:
\[
    \mathbf{x}_{(k + 1)} = \mathcal{O} \bigl( \mathbf{x}_{(k)}, \mathbf{\hat{c}}_{(k)} \bigr)
\]
where $\mathcal{O}$ is a forward operator, coupling the prediction and optimization steps. The final decision is then given by the trajectory endpoint. 
This perspective generalizes Equation \eqref{eq:loss_dfl} by replacing a single optimization problem with a policy induced by the repeated optimization. This formulation naturally extends to our methodology, which we believe is a fitting perspective for understanding how the single-stage optimization in Equation \eqref{eq:pafm_co}, which is isolated at training time, generalizes to the multistage sampling process.

\textbf{Out-of-Distribution Generation.}
One of the strongest arguments for the adoption of constrained sampling algorithms, as opposed to constraint-aware training, is the native ability of sampling-time generalization to arbitrary constraint sets. The evaluation has demonstrated that CAFM effectively handles held-out constraint specifications (Section \ref{exp:pde}) and under-represented design properties (Section \ref{exp:micro}), demonstrating that our training-aligned sampler remains robust to out-of-distribution tasks and constraints. Furthermore, it is worth noting that additional sampling-time constraints can be seamlessly incorporated into CAFM, creating a hybrid between CAFM and PCFM which remains effective due to the sampling-time projections. As such, the proposed approach remains general in its handling of domain shift, avoiding this pitfall of existing constraint-aware training methods.

\textbf{Training Complexity.}
Analogous to decision-focused learning approaches, CAFM's improved downstream performance is provided by incorporating the optimization problem into the training objective. While existing decision-focused learning literature has provided an array of efficiency imporvements that are leveraged by our implementation, the additional computational overhead introduced to solve the projection for each forward pass during training should be acknowledged. To reduce the additional training requirements, the CAFM objective can be applied as a finetuning step, initializing from the pretrained flow matching weights, decreasing the total number of training steps needed to reach convergence.
Appendix \ref{app:warm_start} provides an additional analysis of this finetuning approach to illustrate the efficiency gains provided by warm starting with pretrained weights.



\section{Conclusion}
Motivated by the transformative potential of constraint-aware diffusion and flow matching models for scientific applications, this paper focuses on addressing a fundamental misalignment between the training and sampling process of constrained sampling approaches. To this end, \textit{Constraint-Aware Flow Matching} is proposed, a novel end-to-end framework for constrained generative modeling that directly addresses the mismatch between the learned dynamics and downstream objective. The resulting model is able to generate samples that satisfy constraints while improving data fidelity. The empirical results on three challenging real-world benchmarks demonstrate both the effectiveness and the generality of the proposed approach, reporting state-of-the-art performance across increasingly complex constraint sets and data distributions.
Beyond improving performance in the studied settings, these findings highlight the importance of explicitly accounting for constraint-enforcement mechanisms during training, rather than treating them solely as a post hoc correction at inference time. More broadly, this paper marks a promising step towards building generative models that are both physically consistent and practically reliable in scientific and engineering applications.

\section*{Acknowledgments}
This research is partially supported by NSF awards 2533631, 2401285, 2334936, and by DARPA under Contract No. $\#$HR0011252E005. The authors would like to thank the AiTHENA Program at NASA Langley Research Center for providing partial funding support for this research. The authors acknowledge the Research Computing at the University of Virginia. Any opinions, findings, conclusions, or recommendations expressed in this material are those of the authors and do not necessarily reflect the views of NSF or DARPA.


\bibliography{bib}
\bibliographystyle{unsrtnat}

\clearpage
\appendix

\section{Experimental Setups and Reproducibility}
\label{app:exp_desc}


This section is dedicated to documenting the specific evaluation settings, hyperparameters, and reproducibility details. All experiments were conducted on a single NVIDIA A100-80GB GPU. 
Across all settings, CAFM trains the flow model with the same sampling-time projection mechanism used during constrained inference. Unless otherwise stated, optimization is performed using Adam, with $\beta=(0.9,0.999)$ and no weight decay. 
This is implemented with a differentiable equality-constrained SQP projection layer. Given a predicted endpoint $u$, the projector computes the constraint residual $h(u)$, forms the damped normal matrix $JJ^\top + \epsilon I$, solves for the Lagrange multiplier update, and applies the correction. The default damping is $\epsilon=10^{-6}$, with an adaptive Cholesky fallback. The projected velocity target is computed as
\[
    \frac{u^{\mathrm{proj}}_1 - u_t}{\max(1-t,10^{-3})}.
\]
For projected training losses, we use residual regularization weight $10^{-3}$, sample times $t \in [0,0.95]$, and a projection clamp of $10^{-3}$.

\subsection{Partial Differential Equations}

\paragraph{Evaluation Metrics.}
The selected evaluation criteria follows the metrics used by \citeauthor{utkarsh2025physics} (\citeyear{utkarsh2025physics}). Each is described below:
\begin{itemize}
    \item \textbf{Mean Squared Errors of the Mean (MMSE):} Measures the squared error between the generated mean field and the ground-truth mean field. It evaluates how accurately the model reproduces the expected or average behavior of the target distribution.
    \item \textbf{Standard Deviation (SMSE):} Measures the squared error between the generated standard deviation field and the ground-truth standard deviation field. It evaluates how well the model captures the spatial distribution of uncertainty or variability.
    \item \textbf{Constraint Violation of Boundary / Initial Conditions (CV (BC/IC)):} Measures the constraint violation of the boundary conditions or initial conditions, depending on which are relevant for the problem. Navier--Stokes, Reaction--Diffusion IC, and Burgers IC enforce initial conditions, while Burgers BC enforce boundary conditions.
    \item \textbf{Constraint Violation of Conservation Laws (CV (CL)):} Measures the constraint violation of the conservation laws, which are detemined by the specific problem. These are linear for Navier--Stokes and nonlinear for Burgers and Reaction--Diffusion IC.
\end{itemize}

\paragraph{Baselines.}
The following baselines are adopted for comparison in this experimental setting:
\begin{itemize}
    \item \textbf{Physics-Constrained Flow Matching (PCFM) \cite{utkarsh2025physics}:} Uses an identical sampling setup to CAFM, isolating the impact of end-to-end training.
    \item \textbf{Extrapolation, Correction, and Interpolation (ECI) \cite{cheng2024gradient}:} Uses a similar final state correction to PCFM and CAFM, but operates in a gradient-free manner. Strong performance reported for this PDE setting, making it a relevant baseline.
    \item \textbf{Functional Flow Matching (FFM) \cite{kerrigan2023functional}:} Unconstrained generative flow matching baseline, adapted from standard flow matching for spatiotemporal function spaces.
\end{itemize}

\paragraph{Implementation.}
The PDE experiments evaluate constrained generation across Reaction--Diffusion, Navier--Stokes, Burgers BC, and Burgers IC systems. If not specified, all implementation details follow the implementation from \cite{utkarsh2025physics}. The flow model uses an FNO-based encoder for PDE fields. Unless otherwise specified, the differentiable projection layer uses \texttt{max\_iter=3} SQP/Gauss--Newton correction steps with exact Jacobians.

For the one-dimensional Reaction--Diffusion setting, models are trained for 20,000 iterations with batch size 256 and learning rate $3\times 10^{-5}$. 
The training data is generated following \cite{utkarsh2025physics}; samples have dimensions $128 \times 100$. 
For Navier--Stokes, models are trained for 50,000 iterations with batch size 16 and learning rate $3\times 10^{-5}$. 
Samples have dimensions $32 \times 32 \times 25$. Because of the larger dimensionality of this setting, projection uses \texttt{max\_iter=1} with \texttt{jacobian\_mode=penalty}, \texttt{penalty\_steps=5}, and \texttt{penalty\_step\_size=1e-2}.
For Burgers with boundary-condition constraints, models are trained for 20,000 iterations with batch size 256 and learning rate $3\times 10^{-5}$. 
We use exact Jacobians, \texttt{max\_iter=3}. Samples again have dimensions $101 \times 101$.
For Burgers with initial-condition constraints, models are trained for 20,000 iterations with batch size 256 and learning rate $3\times 10^{-5}$. 
We use exact Jacobians and \texttt{max\_iter=3}. Samples have dimensions $101 \times 101$.

\subsection{Microweather Wind Velocity Field Estimation}

\paragraph{Evaluation Metrics.}
The selected evaluation criteria follows the metrics used by \citeauthor{warner2025generative} (\citeyear{warner2025generative}). Each is described below:
\begin{itemize}
    \item \textbf{Mean Squared Errors of the Mean (MMSE):} Measures the squared error between the generated mean field and the ground-truth mean field. It evaluates how accurately the model reproduces the expected or average behavior of the target distribution.
    \item \textbf{Mean Squared Errors of the Variance (Variance MSE):} Measures the squared error between the generated variance field and the ground-truth variance field. It evaluates how accurately the model reproduces the second-order statistics (variance structure) of the target distribution.
    \item \textbf{Constraint Violation of Coherence Function (CV (Coherence)):} Measures constraint violation of the prescribed coherence function.
\end{itemize}

\paragraph{Baselines.}
The following baselines are adopted for comparison in this experimental setting:
\begin{itemize}
    \item \textbf{Physics-Constrained Flow Matching (PCFM) \cite{utkarsh2025physics}:} Uses an identical sampling setup to CAFM, isolating the impact of end-to-end training.
    \item \textbf{Projected Diffusion Models (PDM) \cite{christopher2024constrained}:} Constrained sampler that applies projections at intermediate steps rather than at the final state. We adapt the methodology to use a flow-matching backbone as opposed to score-based diffusion, keeping the sampler implementation otherwise consistent. 
    \item \textbf{Functional Flow Matching (FFM) \cite{kerrigan2023functional}:} Unconstrained generative flow matching baseline, adapted from standard flow matching for spatiotemporal function spaces.
\end{itemize}

\paragraph{Implementation.}
The wind velocity experiments use a fixed-grid version of the microweather field estimation problem. Unlike the continuous DeepONet representation \cite{lu2019deeponet} used in prior work, this implementation models wind samples directly as fixed-grid tensors \cite{warner2025generative}. Each sample is represented on a $10 \times 10$ spatial grid with a temporal horizon of length 256. Wind values are normalized to the range $[-2,47]$, and conditioning is provided through the sparse SODAR observation pattern.

The flow model is a data-space flow matching network with hidden size 128, four hidden layers, minimum noise scale $\sigma_{\min}=10^{-2}$, and 12 ODE integration steps. The constraint enforces wind coherence over all point pairs. Coherence is computed across the entire grid. The constraint tolerance is $10^{-3}$.
The baseline flow model is trained with batch size 8, 25 epochs, Adam learning rate $10^{-3}$, and gradient clipping at norm 1.0. The CAFM setting is trained with projection enabled, batch size 8, 25 epochs, Adam learning rate $10^{-4}$, and gradient clipping at norm 1.0. Runs are evaluated over seeds 0, 1, and 2.

Training-time projection is performed in data space using an unrolled Augmented Lagrangian method rather than SQP. The training projector and evaluation-time projection use a full-grid augmented Lagrangian method, Adam learning rate $10^{-2}$, \texttt{inner\_iters=64}, \texttt{outer\_iters=8}, \texttt{rho\_init=1.0}, \texttt{rho\_scale=2.0}, \texttt{rho\_max=128.0}, and \texttt{beta=0.0}. Projection is applied at every sampling step.

\subsection{Microstructure Inverse-Design}

\paragraph{Evaluation Metrics.}
\begin{itemize}
    \item \textbf{Mean Squared Error (MSE):} Measures the average squared error between each sample and the ground-truth mean field. This metric is adopted rather than MMSE as the inverse-design targets shift the generated mean field out-of-distribution, making it not directly comparable to the training data. 
    \item \textbf{Nearest Neighbor Mean Squared Error (NNMSE):} Measures the average squared error between each sample and the nearest neighbor in the training set. Benchmarks how close samples are to the ground truth data.
    \item \textbf{Constraint Violation of Porosity (CV (Porosity)):} Measures percentage of samples violating the exact porosity target.
\end{itemize}

\paragraph{Baselines.}
The following baselines are adopted for comparison in this experimental setting:
\begin{itemize}
    \item \textbf{Physics-Constrained Flow Matching (PCFM) \cite{utkarsh2025physics}:} Uses an identical sampling setup to CAFM, isolating the impact of end-to-end training.
    \item \textbf{Projected Diffusion Models (PDM) \cite{christopher2024constrained}:} Constrained sampler that applies projections at intermediate steps rather than at the final state. We adapt the methodology to use a flow-matching backbone as opposed to score-based diffusion, keeping the sampler implementation otherwise consistent. 
    \item \textbf{Flow Matching (FM) \cite{lipman2022flow}:} Unconstrained generative flow matching baseline.
\end{itemize}

\paragraph{Implementation.}
The microstructure experiments evaluate RGB Bentheimer sandstone generation under a target porosity constraint. Images are generated at resolution $256 \times 256$ using the dataset \cite{li2022bentheimer}. During training, images are randomly horizontally flipped and rescaled to $[-1,1]$.

The generative backbone is a UNet with base channel count 128, channel multipliers $[1,1,2,2,4,4]$, two residual blocks per resolution, attention at resolution 16, dropout 0, and EMA decay 0.999. The flow process uses 1000 training flow steps. The baseline flow matching model is trained with batch size 16, Adam learning rate $2\times 10^{-5}$, gradient clipping at norm 1.0, snapshots every 5000 steps, and validation every 250 steps. The CAFM run is trained with batch size 8, maximum 35,000 steps, and Adam learning rate $2\times 10^{-5}$.
Sampling uses Euler integration with 250 steps and batch size 8. Evaluation uses 1000 generated samples. In \texttt{pcfm}, the predicted endpoint is projected at each Euler step. Both \texttt{pcfm} and \texttt{pdm} also apply a hard save-time projection with $\epsilon=0$ to enforce the exact target porosity/count constraint after sampling.

The microstructure constraint is a porosity/count constraint. RGB samples are first converted to grayscale using weights $[0.299,0.587,0.114]$. Pixels are thresholded at 0.0, and the target fraction is set to $k=0.6251$. The projection operator uses an unrolled heuristic top-$k$\slash count projector with a differentiable surrogate update. Projection uses step size 0.1, \texttt{n\_iter=5}, and the fixed-point iteration backpropagation rule. 

\section{Backward-Pass Implementation Details}
\label{app:backward}


Standard unrolling approaches often suffer from high memory cost and backward-pass instability.
Instead, following \citeauthor{kotary2023analyzing} (\citeyear{kotary2023analyzing}), \textit{folded optimization} is adopted, which extends unrolling by differentiating the solver's fixed-point conditions, providing gradients via a single linear solve rather than explicitly unrolling $K$ iterations \cite{kotary2023analyzing}. This enables compatibility with highly optimized blackbox solvers to compute the fixed-point, as it is only necessary to differentiate $\mathbf{x}^{(K)}(\mathbf{\hat{c}})$ rather than unrolling previous iterates.

\paragraph{Efficient Backward-Pass.}
To propagate the gradients through this projection layer, the chain rule is then considered:
\[
    \frac{\partial \mathcal{L}}{\partial v_\theta} = \frac{\partial \mathcal{L}}{\partial \mathcal{P}_C(\hat{z})} \cdot
    \underbrace{\frac{\partial  \mathcal{P}_C(\hat{z})}{\partial \hat{z}}}_{\text{Jacobian of projection}} \cdot \frac{\partial {\hat{z}}}{\partial v_\theta}
\]
However, materializing the Jacobian $\frac{\partial  \mathcal{P}_C(\hat{z})}{\partial \hat{z}}$ explicitly results in a high memory footprint, leading to challenges with scaling to higher dimensional settings and models.
As opposed to explicitly constructing this matrix, consider that the full Jacobian is never required; rather, the vector-Jacobian product (VJP) can be used:
\[
    \Bigl(\frac{\partial  \mathcal{P}_C(\hat{z})}{\partial \hat{z}}\Bigr)^\top u, \quad u = \frac{\partial \mathcal{L}}{\partial \mathcal{P}_C(\hat{z})},
\]
where $u$ is the upstream gradient. As shown by \citeauthor{kotary2023analyzing}, when the projection layer is defined by the fixed-point of an iterative solver, where $z_1 = \mathbf{z}^{\star}(\hat{z}_1) = \Phi(\mathbf{z}^{\star}, \hat{z}_1)$, this differentiation with respect to $\hat{z}_1$ becomes:
\[
    \frac{\partial z_1}{\partial \hat{z}_1} = \frac{\partial \Phi}{\partial \mathbf{z}^{\star}}\frac{\partial z_1}{\partial \hat{z}_1} +  \frac{\partial \Phi}{\partial \hat{z}_1}
\]
The system can be rearranged as:
\[
    \Bigl(I -  \frac{\partial \Phi}{\partial {z}_1^{(k)}} \Bigr) \frac{\partial z_1}{\partial \hat{z}_1} = \frac{\partial \Phi}{\partial \hat{z}_1}
\]
Then, by algebra, the VJP can be computed implicitly by:
\[
    \Bigl(I -  \frac{\partial \Phi}{\partial {z}_1^{(k)}} \Bigr)^\top w = u
\]
and obtain
\[
    \Bigl(\frac{\partial  \mathcal{P}_C(\hat{z})}{\partial \hat{z}}\Bigr)^\top u = \Bigl(\frac{\partial \Phi}{\partial \hat{z}_1} \Bigr)^\top w ,
\]
where $w$ is the solution to the linear system.

\section{Warm Starting Constraint-Aware Training}
\label{app:warm_start}

As mentioned in Section \ref{sec:discussion}, the CAFM objective can be warm started by initializing from pretrained flow matching model weights. While warm starting is not a fundamental requirement of the methodology, we view this as a practical adaptation of CAFM, reducing the training complexity introduced by unrolling the projection operator. 
This is particularly valuable, as including projections in the training forward pass typically results in a \textasciitilde 2.3$\times$ increase in runtime per training step on the PDE suite. 
The subsequent ablation tests how warm starting reduces the number of CAFM training steps needed to reach convergence.

\begin{figure}[ht]
    \centering
    \begin{minipage}[t]{0.72\linewidth}
        \centering
        \includegraphics[width=\linewidth]{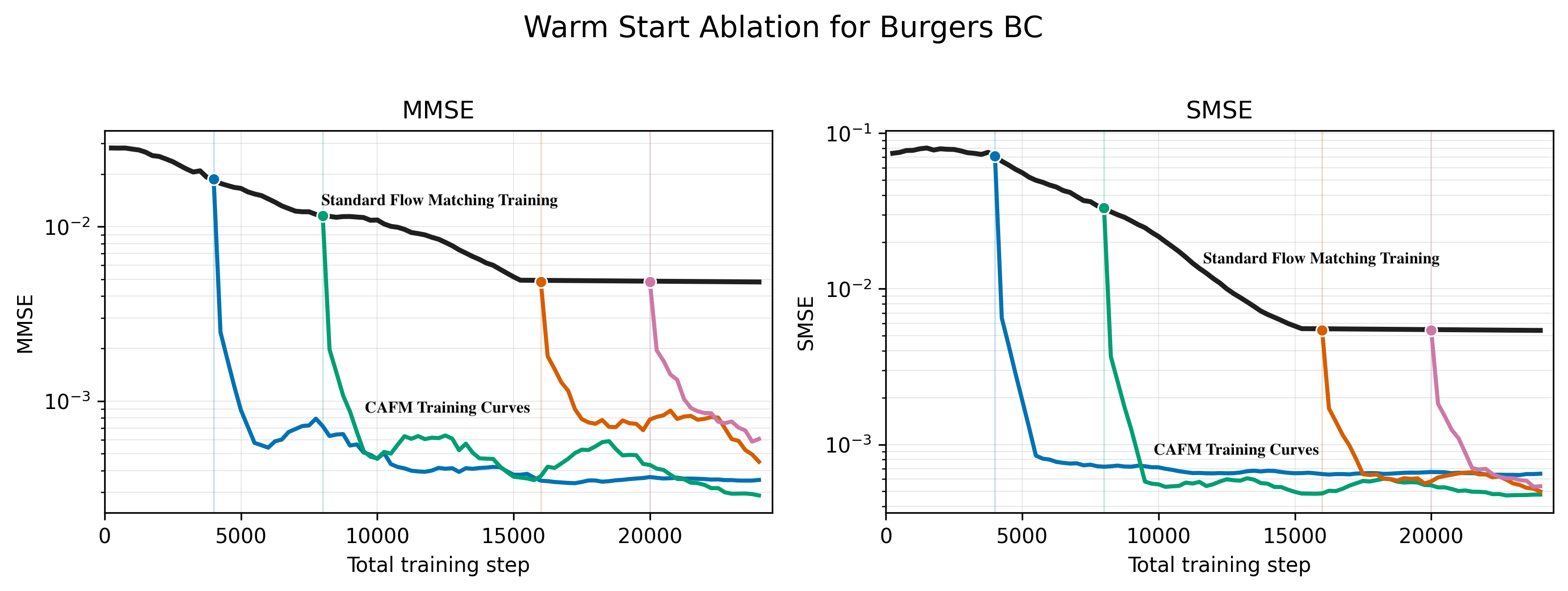}
    \end{minipage}
    \hfill
    \begin{minipage}[t]{0.25\linewidth}
        \centering
        \vspace{-100pt}
        \small
        \renewcommand{\arraystretch}{2.0}
        \setlength{\tabcolsep}{2.5pt}
        \begin{tabular}{lcc}
            \toprule
            & FFM & \shad{CAFM}  \\
            \midrule
            Fwd. & 111.6 ms & \shad{212.4 ms}  \\
            Bwd. & 12.8 ms & \shad{74.6 ms}  \\
            Total & 124.4 ms & \shad{287.0 ms} \\
            \bottomrule
        \end{tabular}

    \end{minipage}
    \caption{\textbf{Left:} Warm starting CAFM training from various points in standard flow matching training on Burgers BC. Additional warm starting steps yield faster convergence of the CAFM objective, with later transitions quickly converging to the same point as earlier transitions. \textbf{Right:} Table details average runtime for forward pass and backward pass operations, reporting the overall runtime impact of CAFM training.}
    \label{fig:warm_start}
\end{figure}

Figure \ref{fig:warm_start} provides empirical evidence that warm starting is useful, with runs adapted to the CAFM objective later in the standard flow matching training process converging more quickly to similar levels of MMSE and SMSE error. This provides strong practical evidence that training complexity can be mitigated through warm starting adaptations, and we defer further exploration of this behavior to subsequent studies.

\section{Unrolling Steps Ablation}
\label{app:unrolling_steps_ablation}

Section~\ref{sec:diff_projection} defines the differentiable projection operator as an iterative optimization layer. In particular, starting from the predicted clean endpoint $\mathbf{x}^{(0)}(\mathbf{\hat{c}})=\hat{z}_1$, the projection is approximated by repeatedly applying an update map
$\Phi$,
\[
    \mathbf{x}^{(k+1)}(\mathbf{\hat c}) = \Phi(\mathbf{x}^{(k)}, \mathbf{\hat c}), \qquad k=0,\ldots,K-1,
\]
so that the projected endpoint used by CAFM is given by
\[
    \mathcal{P}_{C}(\hat z_1) \approx \mathbf{x}^{(K)}(\mathbf{\hat c}).
\]
The number of iterations $K$ dictates the depth of the differentiable projection layer used during training. To study the sensitivity of CAFM to this approximation, we perform an ablation over the number of unrolling steps $K$ in the fixed-grid wind setting.

\begin{table}[h]
  \centering
  \setlength{\tabcolsep}{15.0pt}
  \begin{tabular}{rccc}
    \toprule
    $K$ & MMSE & Variance MSE & CV (Coherence) \\
    \midrule
     1 &   5.515e-02 & 2.981e-01 & 1.65e-03 \\
     2 &  5.479e-02 & 3.109e-01 & \underline{1.20e-04} \\
     4 &  5.472e-02 & 2.947e-01 & \textbf{0.00e+00} \\
     8 &  \textbf{5.288e-02} & \textbf{2.339e-01} & 3.68e-04 \\
    16 & \underline{5.435e-02} & 2.510e-01 & 4.24e-04 \\
    32 & \textbf{5.286e-02} & \underline{2.413e-01} & 5.21e-04 \\
    \bottomrule
  \end{tabular}
  \vspace{10pt}
  \caption{One-seed unrolling-steps ablation for constraint-aware fixed-grid wind sampling.}
  \label{tab:wind-unroll-ablation-seed1}
\end{table}

For each value of $K$, we evaluate the downstream-selected checkpoint trained with that number of projection iterations. During inference, the augmented Lagrangian projection is run with a fixed number of iterations, isolating the results to the training setup. The corresponding results are reported in Table~\ref{tab:wind-unroll-ablation-seed1}.

The ablation suggests that only a few unrolled iterations are necessary for this setting. Increasing the number of projection-layer iterations from $K=1$ to $K=2$ sharply reduces coherence violation, and $K=4$ reaches the training-data floor for this seed under the stronger inference projection. Beyond this point, constraint violation does not improve monotonically, suggesting that the feasibility benefit of additional unrolling steps has largely saturated. Larger values of $K$, however, preserve or slightly improve MMSE and substantially improve Variance MSE as $K \geq 8$.


\end{document}